\pdfoutput=1
\documentclass[11pt]{article}

\usepackage[final]{EMNLP2023}

\usepackage{times}
\usepackage{latexsym}
\usepackage{paralist}
\usepackage{enumerate}
\usepackage{numprint}
\usepackage{xspace}

\usepackage[T1]{fontenc}
\usepackage[utf8]{inputenc}
\usepackage{hyperref}

\usepackage{microtype}
\usepackage{booktabs}
\usepackage{threeparttable}
\usepackage{multirow}
\usepackage{graphicx}
\usepackage{xspace}
\usepackage{tablefootnote}

\def\src{\mathrm{src}}
\def\tgt{\mathrm{tgt}}
\def\hyp{\mathrm{hyp}}
\def\wi{{W\&I+L}\xspace}

\def\fhalf{\ensuremath{\mathrm{F}_{0.5}}\xspace}
\def\fhalfb{\ensuremath{\mathbf{F}_{0.5}}\xspace}

\title{Efficient Grammatical Error Correction Via\\ Multi-Task Training and Optimized Training Schedule}

 \author{
 	Andrey Bout\\
	Huawei Noah's Ark Lab\\
	\texttt{bout.andrey@huawei.com} \\\And
        Alexander Podolskiy\\
	Huawei Noah's Ark Lab\\
	\texttt{podolskiy.alexander} \\ \texttt{@huawei.com}\\\AND
	Sergey Nikolenko\\
        St. Petersburg Department of the \\ Steklov Institute of Mathematics\\
	\texttt{sergey@logic.pdmi.ras.ru}
	\\\And
	Irina Piontkovskaya\\
	Huawei Noah's Ark Lab\\
	\texttt{piontkovskaya.irina} \\ \texttt{@huawei.com}
    } 

\begin{document}
\maketitle
\begin{abstract}
Progress in neural grammatical error correction (GEC) is hindered by the lack of annotated training data. Sufficient amounts of high-quality manually annotated data are not available, so recent research has relied on generating synthetic data, pretraining on it, and then fine-tuning on real datasets; performance gains have been achieved either by ensembling or by using huge pretrained models such as XXL-T5 as the backbone. In this work, we explore an orthogonal direction: how to use available data more efficiently. First, we propose auxiliary tasks that exploit the alignment between the original and corrected sentences, such as predicting a sequence of corrections. We formulate each task as a sequence-to-sequence problem and perform multi-task training. Second, we discover that the order of datasets used for training and even individual instances within a dataset may have important effects on the final performance, so we set out to find the best training schedule. Together, these two ideas lead to significant improvements, producing results that improve state of the art with much smaller models; in particular, we outperform the best models based on T5-XXL (11B parameters) with a BART-based model (400M parameters).
%Progress in neural grammatical error correction (GEC) is hindered by the lack of annotated training data. Sufficient amounts of manually annotated training data are not available, so recent research has relied on generating synthetic data, pre-training on it, and then fine-tuning on real datasets; new performance gains have been achieved by using extremely large pre-trained language models (e.g. XXL-T5) as the backbone. We explore an orthogonal direction: how to use available data more efficiently. Using publicly available tools, we align original and corrected sentences, then we exploit this alignment to create auxiliary sequence-to-sequence tasks. Also, we discover that not only the order of the datasets in training is important but also the order of instances within the same dataset has an effect on the model's performance. A combination of these two ideas leads to improvements, producing results on par with state-of-the-art with much smaller models.
\end{abstract}

\section{Introduction}

Grammatical error correction (GEC) is an important problem setting with obvious applications that often require high-level understanding.
% \paragraph{GEC is treated as a text-to-text problem}
Recent research has developed a common view of grammatical error correction (GEC) as monolingual text-to-text rewriting~\cite{naplava-straka-2019-grammatical,grundkiewicz-etal-2019-neural}. GEC is often tackled with encoder-decoder architectures: the encoder receives an errorful sentence, and the decoder produces its corrected version. In line with mainstream NLP developments, Transformer-based architectures have proven to be beneficial for GEC, leading to recent breakthroughs on multiple benchmarks~\cite{rothe-etal-2021-simple, tarnavskyi-etal-2022-ensembling, sun-wang-2022-adjusting}.
GEC is a challenging task that goes beyond simple spell-checking. The greatest difficulty comes from the complexity of grammar and punctuation rules, inconsistency in rules usage, contextual factors, and multiple correct answers. To solve GEC, a model should have a thorough understanding of grammar, punctuation, and syntax, and be aware of the ambiguity introduced by errors to correctly understand intended meaning. %In the previous research, it was noted that GEC requires learning several independent ``skills''. The model should decide whether a sentence is correct, detect errors in the sentence, choose the best corrections balancing 

Most recent works approach these challenges by providing new complex architectures, using larger models, or ensembling multiple models, e.g. \citet{yuan-etal-2021-multi,rothe-etal-2021-simple,tarnavskyi-etal-2022-ensembling,zhang-etal-2022-syngec}. However, practicality demands faster and smaller models, so in this work, we examine another direction to improve GEC: modifying the training pipeline. Every step is important: \begin{inparaenum}[(i)]
    \item \emph{data preparation} includes generating synthetic data for pretraining and preprocessing available datasets;
    \item \emph{pretraining} is done on synthetic data and/or large-scale datasets with supervised pretext tasks such as language modeling;
    \item \emph{fine-tuning} on downstream datasets is also far from straightforward; at the very least, one has to choose the \emph{order} of training on available datasets, which may significantly impact the results.
    %\item finally, deployment may show that the model is too large and/or slow for the required use case, leading to the need for model distillation or lighter architectures.
\end{inparaenum}
Due to the scarcity of annotated data for fine-tuning, we are concerned with using it as efficiently as possible and introduce two novel approaches for this.

First, one can acquire additional information from parallel annotated corpora, e.g., 
% using, e.g. whether a sentence is correct or not, 
an alignment between two sentences that can be used to derive \emph{operations} that would transform one into the other. One can decompose the editing process into elementary edits: insertion, deletion, replacement, and change of word order; the latter can also be viewed as a sequence of deletions and insertions. This information is exploited by sequence tagging models that show good results on the GEC task, see \citet{omelianchuk-etal-2020-gector,omelianchuk-etal-2021-text,tarnavskyi-etal-2022-ensembling}. A drawback of this approach lies in the need to specify a vocabulary of available operations, use several decoding steps to correct complex errors, and in the hardness/inability to make complex rephrasing. We utilize additional information by converting it into auxiliary tasks and formulating them as sequence-to-sequence problems. As such, the model learns separate ``skills'' required for the successful correction of grammatical errors.

Second, GEC-specific datasets have different quality and sources, e.g., different language proficiency levels. Some datasets are collected from online platforms where users correct each other (they are noisier); in other datasets, sentences are corrected by teachers or annotators (these are cleaner). 
% The former are noisier, while the latter is more clean. 
Some datasets consist of essays written by native speakers; others, of essays written by non-native students. Thus, the distribution of errors may severely differ. It is tempting to remove ``noisy'' or out-of-distribution examples from training sets, but it seems that the model can learn even from such instances. We propose to use a training schedule for GEC datasets and show that it does matter how we order them. Also, we find that the order of sentences within the dataset matters as well; namely, we have found that placing sentences from the same block of text (e.g., the same essay) in the same batch is beneficial. 
 
Our primary contributions are as follows. 
% {\color{red}Update according to the final results}
First, we introduce a multi-task pretraining approach and a fine-tuning schema that together yield an improvement of up to 3\% in \fhalf score compared to similar-sized state of the art models. Second, we show that our approach is able to outperform state of the art \emph{large} models (T5-XL with 3B parameters and T5-XXL with 11B parameters) using a model with 400M parameters, reducing the computational load and ecological footprint. Third, our approach makes the model more robust, improving metrics on several datasets rather than trading them off of each other (as usual). In what follows, Section~\ref{sec:related} surveys related work, Section~\ref{sec:method} introduces our approach, Section~\ref{sec:eval} shows evaluation results, an ablation study, and an analysis of our auxiliary tasks, Section~\ref{sec:concl} concludes the paper, and Section~\ref{sec:limits} discusses the limitations of our approach.

\begin{table}[!t]\centering\setlength{\tabcolsep}{4pt}
\resizebox{\linewidth}{!}{
\begin{tabular}{lrrc}
\toprule
\textbf{Dataset} & \textbf{Sentences} & \textbf{\% errorful} & \textbf{Stages}  \\
\midrule
$\rm C4_{200M}$ & $\rm\sim 180M$ & 99.4 & I \\
PIE-synthetic & $\rm\sim 9M$ & 100.0  & I \\
Lang-8 & \numprint{947344} & 52.5 & II \\
NUCLE & \numprint{56958} & 38.0 & II \\
FCE & \numprint{34490} & 62.4 & II \\
W\&I+L & \numprint{34304} & 67.3 & II, III \\
\midrule
W\&I+L dev & \numprint{4384} & 64.3 & Dev \\
CoNLL test  & \numprint{1312} & 71.9 & Test \\
W\&I+L test  & \numprint{4477} & N/A & Test \\
\bottomrule
\end{tabular}
}
\caption{Dataset statistics and training stages.}\label{table:datasets}
\end{table}

\section{Related work}\label{sec:related}

Neural approaches to grammatical error correction follow two main lines of research:
\begin{inparaenum}[(i)]
\item sequence tagging models and
% that assign transformation tags; %~\cite{omelianchuk-etal-2020-gector};
\item sequence-to-sequence models.
% that follow the general design of machine translation architectures, taking an errorful sentence as input and outputting its corrected version.
\end{inparaenum}
% Below, we discuss prior art on GEC pipelines, emphasizing two main aspects:
% \begin{inparaenum}[(i)]
% \item using synthetic training data and augmentation techniques and
% \item core stages of GEC pipelines.
% \end{inparaenum}

\textbf{Synthetic data}. Training GEC models is difficult due to the natural lack of suitable training data and possible erroneous corrections, so synthetic data becomes a crucial part of any GEC pipeline~\cite{choe-etal-2019-neural,stahlberg-kumar-2021-synthetic,htut-tetreault-2019-unbearable}. It had been used for GEC even before the deep learning era that required larger datasets~\cite{foster-andersen-2009-generrate,brockett-etal-2006-correcting}.
Synthetic data generators can mimic common typos and grammatical errors but usually cannot capture the target error distribution found in real-life evaluation sets or standard benchmarks. Methods for synthetic data generation include character perturbations, dictionary- or edit-distance-based replacements, shuffling word order, rule-based suffix transformations, and more~\cite{grundkiewicz-etal-2019-neural,awasthi-etal-2019-parallel,naplava-straka-2019-grammatical,rothe-etal-2021-simple}. An empirical study of how to generate and use the synthetic data was done in \citet{kiyono-etal-2019-empirical}.

Another line of research 
% applies 
% denoising, augmentation, and data selection 
% strategies to existing datasets to 
upsamples training data in existing datasets. \citet{mita-etal-2020-self} train a GEC model on a natural noisy dataset and then use its outputs for source sentences to construct a less noisy parallel dataset; \citet{zhang2019sequence} use sentence rewriting approaches;
% to craft data for the fine-tuning stage. 
\citet{lichtarge-etal-2020-data} apply delta-log-perplexity scoring to rank sentences according to the difference in perplexity between two base model checkpoints and use higher-scoring sentences for final fine-tuning. 
 
\textbf{Multi-stage fine-tuning}.
Due to data scarcity, training GEC models from scratch could be cumbersome. One of the options is to pre-train a model on some auxiliary task, e.g. \citet{choe-etal-2019-neural} proposed to initialize the GEC model with the pre-trained denoising autoencoder. Many GEC pipelines utilize pre-trained language models as backbone models for GEC; in particular, \citet{rothe-etal-2021-simple} used T5~\cite{raffel2020exploring} while~\citet{katsumata-komachi-2020-stronger} and~\citet{sun-wang-2022-adjusting} used BART~\cite{lewis-etal-2020-bart}. Pretrained language models are also beneficial for reranking output hypotheses generated with beam search~\cite{kaneko-etal-2019-tmu}. \citet{choe-etal-2019-neural}, \citet{omelianchuk-etal-2020-gector} and \citet{tarnavskyi-etal-2022-ensembling} decompose the training process of a GEC model into several stages:
\begin{inparaenum}[(i)]
    \item pre-training on an errorful synthetic dataset;
    \item fine-tuning on natural high-quality datasets that combine both errorful and error-free sentences.
\end{inparaenum}
Each stage requires its own tuning of hyperparameters
such as the number of training steps and the learning rate. 

\textbf{Multi-task learning}.
Several works aim to utilize additional information along with the standard parallel mapping. First, the grammatical error detection (GED) task can be extracted from GEC; \citet{yuan-etal-2019-neural} perform multi-task training with GED and GEC tasks and use GED features for reranking. A similar approach by \citet{fang-etal-2020-hybrid} trained a GED model separately and used it to filter edits. \citet{yuan-etal-2021-multi} separately pretrain a GED model and use its outputs as auxiliary inputs to fine-tune the encoder-decoder GEC model and rerank its outputs. \citet{zhang-etal-2022-syngec} incorporate syntactic dependency information into the encoder.

\textbf{Non-autoregressive decoding}.
Another line of research introduces non-autoregressive decoding to speed up models. \citet{Awasthi2019ParallelIE} predict language-specific edits to be applied to the output sequence. Iterative refinement is also possible. \citet{Gu2019LevenshteinT} non-autoregressively refine an output sequence using language-agnostic insertions and deletions. \citet{yakovlev-etal-2023-gec} decompose the inference stage into permutation and decoding. First, a permutation network repositions the tokens of an input sequence with possible deletions and insertions. Second, the intermediate sequence is passed to a decoder network that iteratively fills in inserted placeholders.

\textbf{GPT-3.5 and GPT-4}. 
The recent success of GPT-based models for a wide variety of tasks has led to several parallel works that compare how well these models fare in grammatical error correction. \citet{fang2023chatgpt} show that ChatGPT is still worse on GEC benchmarks than fine-tuned sequence-to-sequence models, both in the zero-shot and few-shot scenarios. \citet{coyne2023analyzing} provide the corresponding analysis for GPT-4, with even lower results, which means that specialized models are still relevant for GEC scenarios and validate our research.

\section{Approach}\label{sec:method}
In this section, we introduce our approach that uses multi-task learning and optimizes the training schedule for the sequence-to-sequence model architecture. In Section~\ref{subsection:model}, we outline the model and hyperparameters used for training.  In Section~\ref{subsection:datasets}, we describe existing GEC datasets for training and evaluation, highlighting the specific properties of each. In Section~\ref{subsection:multitask}, we specify the main and auxiliary tasks formulated as sequence-to-sequence mapping. Section~\ref{subsection:stages} describes the training steps.

\subsection{Model}\label{subsection:model}
We use a straightforward text-only approach while also keeping the model size limited. Namely, we formulate all training tasks as sequence-to-sequence text rewriting problems and do not introduce any additional heads for edit prediction or other tasks.
As the backbone, we use the BART~\cite{lewis-etal-2020-bart} model with a 12-layer encoder and decoder.  We train the model in fp16 mode with learning rate $1e^{-5}$ and warmup ratio 7\% with a linear scheduler. We fit 12 instances in a batch and use gradient accumulation with 4 steps to achieve a larger effective batch size. We did not perform a hyperparameter search except for the learning rate, which was chosen from [$1e^{-5}$, $5e^{-5}$, $1e^{-6}$]. All training experiments were done on 8 NVIDIA Tesla V100 GPUs with 32GB of memory.

%% Part about the data being used
\subsection{Training Data}\label{subsection:datasets}
%\textbf{[TODO: in this part, we need to describe the properties of the datasets in human-understandable way: describe sources of the natural data, specific features of each dataset and the core difference between synthetic and natural data. Then, in the whole Approach section, we should avoid dataset names, use their characterisics instead: "synthetic", "errorful", "esseys", "the most clean" etc]} 

\begin{figure*}[!t]\centering
\includegraphics[width=\linewidth]{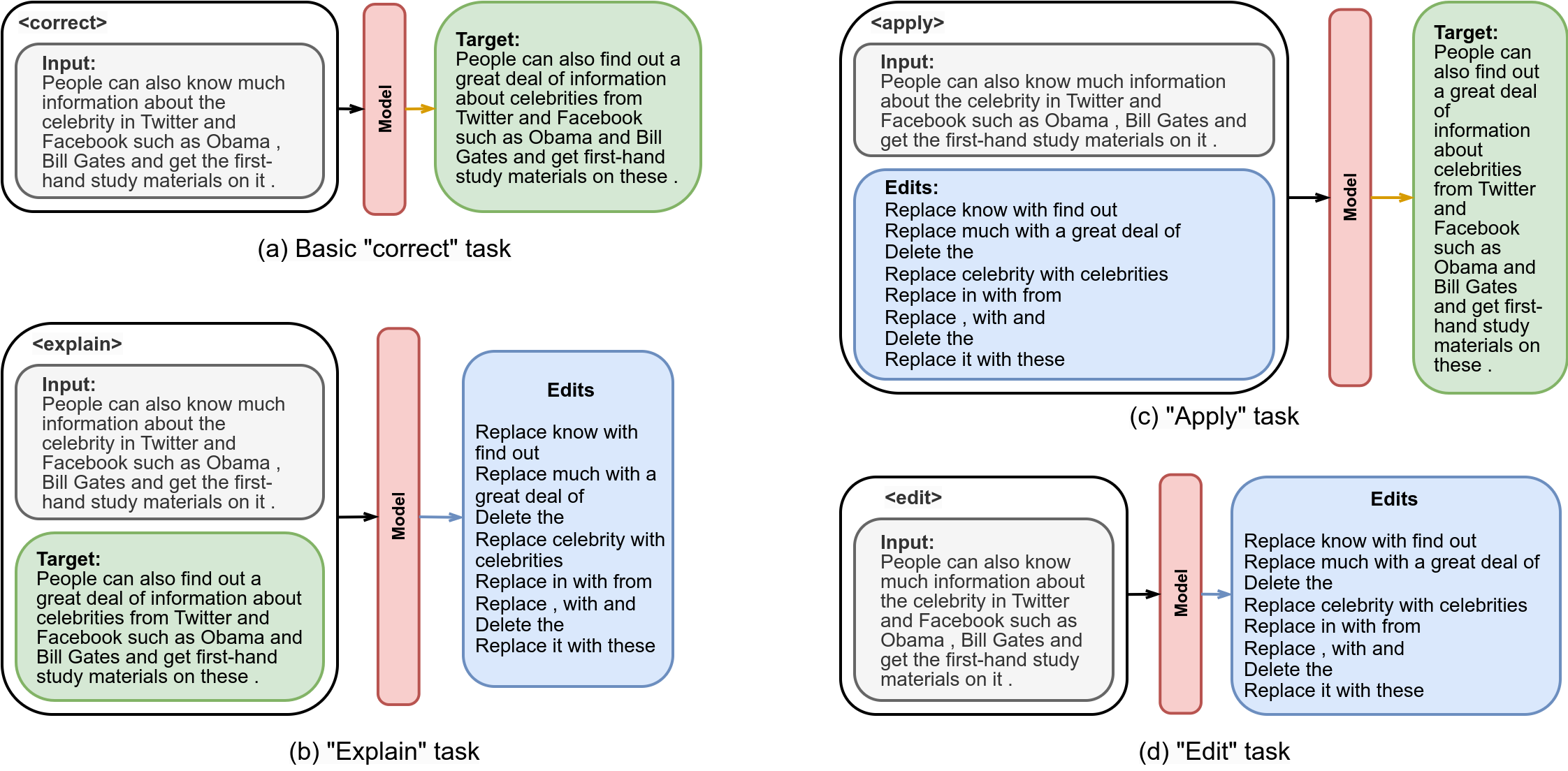}

\caption{Tasks automatically generated from a source-target pair for multi-task pretraining.}\label{fig:mt}
\end{figure*}

For pretraining (Stage I), we use $\rm C4_{200M}$ or PIE datasets. $\rm C4_{200M}$ is a synthetic corpus based on clean sentences from the C4 dataset. This corpus was generated by a tagged corruption model to meet the error distribution of BEA-dev~\cite{bryant-etal-2019-bea}; see~\citet{stahlberg-kumar-2021-synthetic} for details. PIE is a synthetic dataset of 9M parallel sentences generated by using rule-based grammatical errors such as deletion, insertion, and word replacement~\cite{awasthi-etal-2019-parallel}.

For other training stages, we use the following datasets:
\begin{inparaenum}[(i)]
\item National University of Singapore Corpus of Learner English (NUCLE)~\cite{dahlmeier-etal-2013-building} that consists of essays written by undergraduate students on different topics and annotated by professional English instructors;
\item Lang-8 Corpus of Learner English (Lang-8)~\cite{tajiri-etal-2012-tense} collected from the online language learning site Lang-8; this dataset is relatively ``noisy'' as the users corrected themselves, and it comes with several annotations;
\item First Certificate in English (FCE)~\cite{yannakoudakis-etal-2011-new} with short texts written by English learners as answers to exam questions assessing the upper-intermediate level; this dataset is relatively clean but covers only a single group of English learners;
\item Write \& Improve + LOCNESS Corpus (\wi)~\cite{bryant-etal-2019-bea}; Write \& Improve dataset consists of text blocks (essays, letters etc.) written by English learners and submitted to the W\&I system; LOCNESS is composed of essays written by native English students, used only for evaluation; these datasets are the ``cleanest'' and have a wide distribution over different levels of English.
\end{inparaenum}
Here and below, \emph{errorful} sentences are those that contain at least one error.
We use the BEA-2019 development set, i.e. \wi-dev, to choose the best model and report results on the CoNLL2014 test set~\cite{ng-etal-2014-conll} evaluated by the official M2 scorer~\cite{dahlmeier-ng-2012-better} and the BEA-2019 test set evaluated by ERRANT~\cite{bryant-etal-2017-automatic}.
Table~\ref{table:datasets} summarizes dataset statistics and shows which training stages they are used for.

\subsection{Multi-task learning} \label{subsection:multitask}

The standard approach is to view GEC as a sequence-to-sequence mapping from errorful sentences to corrected ones. Since the sentences are from the same language, one can align two sentences using an edit distance algorithm. The alignment can be transformed into a sequence of \textit{insert}, \textit{delete}, and \textit{replace} operations that transform the errorful sentence into a correct one. We find the list of operations using ERRANT\footnote{We use version 2.3 of ERRANT for both evaluation and training set construction.}~\cite{Bryant2019TheBS} and use them for auxiliary tasks. 

Inspired by recent works on chain of thought prompting~\cite{wei2023chainofthought, zhou2022least}, we construct several tasks and examine their influence on the model's performance. Each task, as illustrated in Fig.~\ref{fig:mt}, is explicitly marked by a prefix written as $\langle\textit{prefix}\rangle$ and followed by an input string:
\begin{enumerate}[(i)]
    \item \textbf{Correct}: standard generation of a corrected sentence given the original, encoded as ``\emph{$\langle\textit{correct}\rangle$ source}'' and trained to produce the target sentence (Fig.~\ref{fig:mt}a);
    \item \textbf{Explain}: generation of a sequence of explanations (atomic edits) given both original and corrected sentences; encoded as ``\emph{$\langle\textit{explain}\rangle$ Input: src \textbackslash n Target: tgt}'' with the result in the format ``\emph{Delete smth \textbackslash n Insert smth...}'' (Fig.~\ref{fig:mt}b); if no correction is needed, the model should generate ``\emph{No correction}'';
    \item \textbf{Apply}: application of edits to the original sentence to get the target, encoded as ``\emph{$\langle\textit{apply}\rangle$ Input: src \textbackslash n Do: Delete smth \textbackslash n Insert smth \textbackslash n Replace smth with smth}''; the result should be the correct target sentence (Fig.~\ref{fig:mt}c);
    \item \textbf{Edit}: generation of the sequence of edits given the original sentence but not the target, encoded as ``\emph{$\langle\textit{edit}\rangle$ src}'' with the target in the form ``\emph{Delete smth \textbackslash n Insert smth \textbackslash n Replace smth with smth}'' (Fig.~\ref{fig:mt}d); if no correction is needed, the model should generate ``No correction''.
\end{enumerate}

We generate an auxiliary sequence-to-sequence pair for every task and for every pair from the original dataset.
% Chain-of-thought like prompts show substantially lower performance.
%Our approach to multi-task learning follows recent works on Chain-of-thought prompting, where the model learns not only predict the final result, but also generates intermediate reasoning steps explaining how this result was obtained. Especially, we are inspired by the work \textbf{[TODO: references]}, where the authors show that using explanations as an additional task during the training increases the performance on the base task without increasing the final inference time. 

%We propose to use step-by-step explanation format, presented in Table {} \textbf{[TODO: table with task format example]}. We decompose the correction process into \textit{insert}, \textit{delete} and \textit{replace} operations, and, given a pair of source (erroneous) and target (corrected) sentences, supply it with an explanation, consisting of the sequence of the required operations, or ``No correction'', if the source is equal to the target. Then, we construct three tasks, which we than train simultaneously, mixing them in each batch: 

\subsection{Training Order} \label{subsection:stages}

\begin{figure}[!t]\centering
\includegraphics[width=\linewidth]{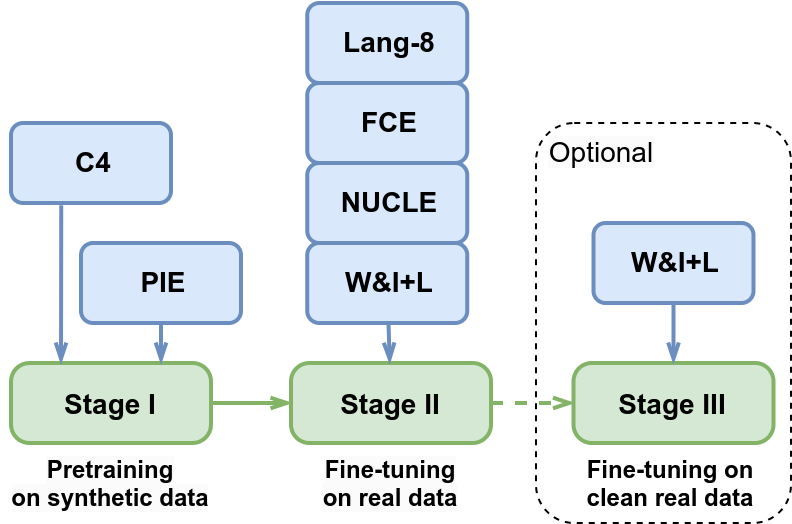}

\caption{Training order for a GEC model.}\label{fig:order}
\end{figure}

\citet{tarnavskyi-etal-2022-ensembling} and \citet{omelianchuk-etal-2021-text} mention that training order is essential to obtain the best possible results. They separate the overall process into three stages: pretrain the model on synthetic data, fine-tune it on errorful sentences from the four GEC datasets: LANG-8, NUCLE, FCE, and W\&I+L, and then fine-tune it on the clean GEC dataset W\&I+L. 
%We take this idea further in our five-stage training pipeline (Fig.~\ref{fig:pipeline}).
% detailed below.

In this work, we suggest to modify this procedure and claim that the training process benefits from choosing a correct ordering of data. Our process is illustrated in Fig.~\ref{fig:order}: on Stage I, we similarly pretrain our model on large-scale synthetic data. We consider two datasets that differ both in size and in generation approach: $\rm C4_{200M}$ and PIE. This step adapts the model for the downstream task.
On Stage II, we fine-tune on four GEC datasets but modify the above procedure. First, we use all sentences, not only errorful ones. Second, we use the datasets in a strict order: Lang-8, NUCLE, FCE, W\&I+L, with no shuffling across datasets. Third, we do not shuffle samples inside the datasets either: each dataset is composed of coherent texts, so we thus preserve their original structure and place examples from the same texts together.

For a more complete comparison with previous works, we add Stage III where we additionally fine-tune the model on the W\&I+L dataset.
%to get improved results on the BEA-dev. 
We note that this step helps to increase recall without a substantial decrease in precision, yielding modest improvements in \fhalf. Note that in previous works, this step was mandatory as the target distribution is correlated with the W\&I+L dataset. In contrast, we add this dataset as the last in our training order, which is a more natural approach; the suggested scheme also looks more suitable for real world tasks where there is no obvious way to split the data into different stages.

%\subsection{Precision-Recall Tradeoff } \label{subsection:tradeoff}

\subsection{Re-weighted Sampling} \label{subsection:sampling}

The \fhalf metric is commonly used for the evaluation of GEC models. It puts a higher weight on the model's precision than recall. In our experiments, we see that the proposed training pipeline makes the model less aggressive in editing which improves performance.
Another approach to making the model choose corrections that it is confident about is the Align-Pred method proposed by \citet{sun-wang-2022-adjusting}. It increases the probability of the next original token (obtained by aligning the original and currently generated sequences) during decoding. In order to show that our method is orthogonal to this, we apply Align-Pred to our best model. We introduce a modification to this method that goes beyond the original, applying temperature scaling before Align-Pred. This significantly improves Align-Pred (see Table~\ref{table:results}).

\begin{table*}[!t]
\centering\setlength{\tabcolsep}{3pt}
\resizebox{\linewidth}{!}{
\begin{tabular}{lrc|ccc|ccc}\toprule
& & & \multicolumn{3}{c|}{\textbf{CoNLL-14}} & \multicolumn{3}{c}{\textbf{BEA-test}} \\
\textbf{Model} & \textbf{Size} & \textbf{Synthetic} & \textbf{Prec} & \textbf{Rec} & \fhalfb & \textbf{Prec} & \textbf{Rec} & \fhalfb \\\midrule
%& Our fine-tune full& 70.58 & \textbf{53.35} & 66.30 \\
%& Our filtered full & 72.29 & 52.79 & 67.32 \\
T5~\cite{rothe-etal-2021-simple} & 770M & mC4(>300M)+CLANG & --- & --- & 66.04 &  --- & --- & 72.06 \\
T5-XL~\cite{rothe-etal-2021-simple} & 3B & mC4(>300M)+CLANG &  --- & --- & 67.65${}^\ast$ &  --- & --- & 73.92 \\
T5-XXL~\cite{rothe-etal-2021-simple} & 11B & mC4(>300M)+CLANG &  --- & --- & 68.75${}^\ast$ & --- & --- & \underline{75.88} \\
GECToR-XLNet~\cite{omelianchuk-etal-2020-gector} & 350M & PIE(9M) & 77.5 & 40.1 & 65.3 & 79.2 & 53.9 & 72.4 \\
GECToR-RoBERTa~\cite{tarnavskyi-etal-2022-ensembling} & 350M & PIE(9M) & 74.40 & 41.05 & 64.0 & 80.70 & 53.39 & 73.21 \\
GECToR-XLNet~\cite{lai-etal-2022-type} & 350M & PIE(9M) & \textbf{78.18} & 42.67 & 67.02 & \textbf{81.33} & 51.55 & 72.91 \\
Transformer~\cite{stahlberg-kumar-2021-synthetic} & 340M & $\mathrm{C4_{200M}}$ & 72.8 & \underline{49.5} & 66.6 & 72.1 & 64.4 & 70.4 \\
BART-based~\cite{sun-wang-2022-adjusting} & 400M & 300M & --- & --- & --- & 76.1 & \textbf{65.6} & 73.8 \\
BART-based + Align-and-Predict~\cite{sun-wang-2022-adjusting} & 400M & 300M & --- & --- & --- & 78.7 & 63.2 & 75.0 \\
\midrule
BART (ours) & 400M & $\rm C4_{200M}$ & 75.43 & \textbf{51.20} & \underline{68.91} & 78.19 & \underline{65.54} & 75.28 \\
BART (ours) + Align-and-Predict + Temp. Scaling & 400M & PIE(9M) & 75.45 & 47.03 & 67.31 & 78.49 & 58.66 & 73.52 \\
BART (ours) + Align-and-Predict + Temp. Scaling & 400M & $\rm C4_{200M}$ & \underline{78.00} & 49.12 & \textbf{69.79} & \underline{80.88} & 61.15 & \textbf{75.97} \\
\bottomrule
\end{tabular}
}
\caption{Evaluation results on CoNLL-14 and BEA-test. Best results are shown in \textbf{bold}, second best are \underline{underlined}; ${}^\ast$ results shown by the \emph{arXiv} version~\cite{rothe2022arxiv} that differs from~\cite{rothe-etal-2021-simple} on the CONLL test set}\label{table:results}
\end{table*}

\begin{table}[!t]
\centering\setlength{\tabcolsep}{4pt}
% \begin{threeparttable}
\resizebox{\linewidth}{!}{
\begin{tabular}{p{.18\linewidth}|p{.20\linewidth}|p{.20\linewidth}|rrr}
\toprule
% \begin{tabular}[c]{@{}l@{}}Pretraining\\ Dataset\end{tabular} & 
\textbf{Pre\-train\-ing} &
\textbf{Stage II} & \textbf{Stage III} & \textbf{Prec} & \textbf{Rec} & \fhalfb \\
\midrule
$\rm C4_{200M}$     & ---        & ---          &  36.67 & 24.83  & 33.88  \\
$\rm C4_{200M}$     & 4 datasets        & W\&I+L       & \textbf{66.77}  & \textbf{50.14}  & \textbf{62.62}  \\
\midrule
PIE   & ---        & ---          &  44.55  & 20.04   &  35.79 \\
PIE & 4 datasets     & W\&I+L      & 65.00
  & 46.29  & 60.14 \\
\bottomrule
\end{tabular}}
% \end{threeparttable}
\caption{Influence of the pretraining dataset. Evaluation results on BEA-dev.}
\label{tab: pret compar}
\end{table}

\begin{table*}
\centering\setlength{\tabcolsep}{8pt}
\resizebox{\linewidth}{!}{
\begin{tabular}{l|ccc|ccc|ccc}\toprule
& \multicolumn{3}{c|}{\textbf{CoNLL-14}} & \multicolumn{3}{c|}{\textbf{BEA-test}} &  \multicolumn{3}{c}{\textbf{BEA-dev}}\\
\textbf{Training approach} & \textbf{Prec} & \textbf{Rec} & \fhalfb & \textbf{Prec} & \textbf{Rec} & \fhalfb & \textbf{Prec} & \textbf{Rec} & \fhalfb \\\midrule
Two stages & 66.47 & 47.65 & 61.60 & 59.61 & 63.74 & 60.39 & 47.94 & 40.91 & 46.35 \\
Three stages & 73.64 & \underline{53.16} & 68.37 & 75.65 & \textbf{68.94} & 74.20 & 66.77 & \underline{50.14} & \underline{62.62} \\
Two stages, optimized schedule & \textbf{77.18} & 48.52 & \textbf{69.02} & \textbf{78.71} & 63.05 & \underline{74.90} & \underline{68.60} & 41.56 & 60.70 \\
Three stages, optimized schedule & 74.96 & 52.34 & \underline{69.00} & 77.17 & 66.39 & 74.74 & 68.15 & 46.31 & 62.28 \\
%0.6662	0.4693	0.6146
%0.6651	0.4747	0.6158 <- next iteration Need more training steps 
Multi-task: three tasks, two stages & 66.51 & 48.47 & 61.79 & 59.55 & 63.53 & 60.31 & 48.22 & 41.50 & 46.71 \\
% Maybe here need more training steps too
Multi-task: three tasks, three stages & 73.96 & \textbf{53.30} & 68.64 & 75.85 & \underline{68.63} & 74.29 & 65.62 & \textbf{52.07} & 62.38 \\
Multi-task: three tasks, optimized schedule & \underline{75.43} & 51.20 & 68.91 & \underline{78.19} & 65.54 & \textbf{75.28} & \textbf{68.64} & 46.50 & \textbf{62.67} \\
\bottomrule
\end{tabular}
}
\caption{Comparison of standard and multi-task fine-tuning; all models are pretrained on $\rm C4_{200M}$. Best results are shown in \textbf{bold}, second best are \underline{underlined}.}\label{table:ablation}
\end{table*}

\section{Evaluation and Analysis}\label{sec:eval}

\subsection{Comparison with baselines} We compare the proposed model with several state of the art baselines. We used three variations of the tagging model GECToR: with the XLNet-L backbone~\cite{omelianchuk-etal-2020-gector}, with RoBERTa backbone and multi-stage training~\cite{tarnavskyi-etal-2022-ensembling}, and with the XLNet-L backbone and multiturn corrections \cite{lai-etal-2022-type}. For \emph{seq2seq} baselines we consider the vanilla GEC Transformer trained on $\rm C4_{200M}$~\cite{stahlberg-kumar-2021-synthetic}, T5-XL and T5-XXL models~\cite{rothe-etal-2021-simple} also trained on the C4 dataset, and a BART-based model trained on 300M synthetic data~\cite{sun-wang-2022-adjusting}. All baselines have comparable or larger model sizes than ours and use the same PIE synthetic dataset or larger synthetic collections.

Table~\ref{table:results} shows evaluation results for the CoNLL-14~\cite{ng-etal-2014-conll} and BEA-test~\cite{bryant-etal-2019-bea} datasets. Our approach shows the best results for comparable model sizes with a significant margin and outperforms all models, including T5-XL with 3B parameters and even T5-XXL which is 30x larger than our model (11B parameters) and trains on a larger synthetic dataset. We also present evaluation results of our approach with pretraining on the PIE dataset to compare with the models pretrained on it. Again, we see that our model outperforms more sophisticated methods such as~\cite{lai-etal-2022-type}.

% The small gap in Bea-dev results (62.67 vs 62.62) appears to arise from the difference in error distributions across train, dev, and test set; this is our hypothesis for BEA-test as far as this set is unavailable for analysis (we cannot attach the picture, but we add it to the final version of paper). Using the suggested combination of optimized schedule and multi-task learning we find better precision-recall tradeoffs for more difficult test sets. In this case, the precision with MTL and an optimized schedule is much better than with a three-stage approach. We did not explicitly mention it in the paper but we are going to add it to the final version if accepted.

\subsection{Influence of the Pretraining Dataset}

In this section, we analyze the choice of the pretraining dataset,
% We are concerned with the size and the type of generated errors. 
comparing two publicly available synthetic datasets: $\rm C4_{200M}$ and PIE. They differ not only in size (see Table~\ref{table:datasets}) but also in the type of generated errors that are model-based and rule-based respectively. Table~\ref{tab: pret compar} shows the performance of models pretrained in different setups. The model pretrained on $\rm C4_{200M}$ has higher recall indicating that it covers more errors, while the model pretrained on PIE reaches higher precision. Note that almost all sentences from synthetic datasets contain errors, which means that the pretraining distribution differs a lot from the development set. Hence, to make a fair comparison we further fine-tune the models on GEC-specific datasets using the standard three-stage approach. Table~\ref{tab: pret compar} shows that the model pretrained on $\rm C4_{200M}$ performs better in terms of precision, recall, and the \fhalf score.

\subsection{Order of the Datasets}

\begin{table}[!t]
\centering\setlength{\tabcolsep}{3pt}
% \begin{threeparttable}
\resizebox{\linewidth}{!}{
\begin{tabular}{l|rrr}
\toprule
\textbf{Order} & \textbf{Prec} & \textbf{Rec} & \fhalfb \\
\midrule
\multicolumn{4}{c}{\textbf{Order preserved within each dataset}} \\
\midrule
FCE $\rightarrow$ Lang-8 $\rightarrow$ NUCLE $\rightarrow$ W\&I+L       & 68.72 & 44.98 & 62.15 \\
NUCLE $\rightarrow$ Lang-8 $\rightarrow$ FCE $\rightarrow$ W\&I+L       &   68.86  & 45.02  &  62.26   \\
Lang-8 $\rightarrow$ FCE $\rightarrow$ NUCLE $\rightarrow$ W\&I+L       &  68.52 &  45.32  &  62.16  \\
Lang-8 $\rightarrow$ NUCLE $\rightarrow$ FCE $\rightarrow$ W\&I+L       &  68.78  &  45.98 &  62.67 \\
FCE $\rightarrow$ Lang-8 $\rightarrow$ W\&I+L $\rightarrow$ NUCLE        &   77.41  & 13.70  & 40.11 \\
\midrule
\multicolumn{4}{c}{\textbf{Instances shuffled within each dataset}} \\
\midrule
FCE $\rightarrow$ Lang-8 $\rightarrow$ NUCLE $\rightarrow$ W\&I+L      & 66.41 &    48.47 & 61.84 \\
Lang-8 $\rightarrow$ NUCLE $\rightarrow$ FCE $\rightarrow$ W\&I+L      & 66.99 &    48.48 &  62.24     \\
\midrule
\multicolumn{4}{c}{\textbf{Instances shuffled across datasets}} \\
\midrule
Lang-8, NUCLE, FCE, W\&I+L & 48.21 & 41.20 & 46.63 \\
\bottomrule
\end{tabular} }
% \end{threeparttable}
\caption{Ablation study of the dataset fine-tuning order. The model is trained in 2-stages, evaluated on BEA-dev.}
\label{tab:dataset order 2-stage}
\end{table}

\begin{table}[!t]
\centering\setlength{\tabcolsep}{3pt}
% \begin{threeparttable}
\resizebox{\linewidth}{!}{
\begin{tabular}{l|c|rrr}
\toprule
\textbf{Stage II} & \textbf{Stage III} & \textbf{Prec} & \textbf{Rec} & \fhalfb \\
\midrule
\multicolumn{5}{c}{\textbf{Order preserved within each dataset}} \\
\midrule
FCE $\rightarrow$ Lang-8 $\rightarrow$ NUCLE $\rightarrow$ W\&I+L    & W\&I+L       &   67.88  &  47.02 &	62.34 \\
NUCLE $\rightarrow$ Lang-8 $\rightarrow$ FCE $\rightarrow$ W\&I+L    & W\&I+L       & 68.23  &  47.12 &  62.62 \\
Lang-8 $\rightarrow$ FCE $\rightarrow$ NUCLE $\rightarrow$ W\&I+L   &  W\&I+L        &  68.14  &  47.61 &	62.73 \\
Lang-8 $\rightarrow$ NUCLE $\rightarrow$ FCE $\rightarrow$ W\&I+L    & W\&I+L       & 67.35  &   49.89 &  62.94 \\
FCE $\rightarrow$ Lang-8 $\rightarrow$ W\&I+L $\rightarrow$ NUCLE   &   W\&I+L      &  67.66   &  47.32 & 62.30 \\
\midrule
\multicolumn{5}{c}{\textbf{Instances shuffled within each dataset}} \\
\midrule
FCE $\rightarrow$ Lang-8 $\rightarrow$ NUCLE $\rightarrow$ W\&I+L & W\&I+L & 66.05 &	50.03 &	62.08 \\
Lang-8 $\rightarrow$ NUCLE $\rightarrow$ FCE $\rightarrow$ W\&I+L & W\&I+L   & 66.05 &    50.34 &  62.17   \\
\midrule
\multicolumn{5}{c}{\textbf{Instances shuffled across datasets}} \\
\midrule
Lang-8, NUCLE, FCE, W\&I+L & W\&I+L & 65.62 & 52.07 & 62.38 \\
\bottomrule
\end{tabular} }
% \end{threeparttable}
\caption{Ablation study of the dataset training order. The model is trained in 3-stages. Evaluation is done on the BEA-dev dataset.}
\label{tab:dataset order 3-stage}
\end{table}

We also examine the influence of the ordering of training datasets and examples within a dataset. We are mainly concerned with multi-task training, but note that the training schedule also impacts the single-task pipeline (see Table~\ref{table:ablation}). The model trained using a specific schedule achieves good results after the first stage of fine-tuning, while the third stage improves the model's recall and makes its results surpass the standard three-stage training.

For the multi-task approach, we use three tasks---Correct, Explain, and Apply---and different dataset schedules. There are 16 possible orderings, but Lang-8 is a ``noisy'' dataset so it is reasonable to place it early in the fine-tuning, while W\&I+L is ``cleaner'' than others so we use it late in the training. Tables~\ref{tab:dataset order 2-stage} and \ref{tab:dataset order 3-stage} shows the performance of models trained in different setups. It shows that, first, the order of datasets really matters: some cases outperform others by a significant margin. Second, shuffling instances within a dataset also reduces performance, which supports our hypothesis that sentences from the same block of text should be placed together while training. Third, the best performance is achieved when W\&I+L is the last dataset in the schedule, as evidenced by the drop in performance for the ``FCE~$\rightarrow$ Lang-8 $\rightarrow$ W\&I+L $\rightarrow$ NUCLE'' setup and improvement after further fine-tuning on the W\&I+L dataset. %Third, the third stage improves the model's recall without a substantial drop in the precision, except for the Lang-8 - FCE - NUCLE - W\&I+L schedule.

\subsection{Multi-Task Pretraining}

\begin{table}[!t]
\centering
% \begin{threeparttable}
\resizebox{\linewidth}{!}{
\begin{tabular}{l|r|r|r}
    \toprule
    \textbf{Tasks}  &  \textbf{Prec}  &  \textbf{Rec}  &  \fhalfb \\
    \midrule
    Correct + Explain                              &  69.03  &  44.20  &   62.06   \\
    Correct + Apply                            &  68.64  &  \textbf{46.50}  &   \textbf{62.67}   \\
    Correct + Edit                             &  \textbf{69.19}  &  44.92  &   62.45   \\
    Correct + Explain + Apply                      &  68.78  &  45.98  &   62.57   \\
    Correct + Apply + Edit                     &  68.98  &  44.95  &   62.32   \\
    Correct + Explain + Apply + Edit               &  68.58  &  45.81  &   62.38   \\
    Edit + Apply                            &  49.41  &  28.80  &   43.22   \\
    \bottomrule
\end{tabular} }
% \end{threeparttable}
\caption{Ablation study on auxiliary tasks. Evaluation is done on the BEA-dev dataset.}
\label{tab:task abl}
\end{table}

Table~\ref{table:ablation} compares standard fine-tuning with the proposed multi-task fine-tuning. We show the commonly used metrics---precision, recall, and \fhalf score---on three GEC datasets: CoNLL-14, BEA-test, and BEA-dev. In all cases, multi-task training leads to improved performance for the final task, especially for the BEA-dev dataset.
 
Table~\ref{tab:task abl} presents an ablation study related to the choice of auxiliary tasks; we fix the training schedule and use different task compositions. We make the following observations. First, the best combination of tasks is ``Correct'' and ``Apply'', yielding the highest recall and high enough precision compared to other settings. Second, adding the edit prediction task (``Edit'') in combination with other tasks lowers the model's performance; below, we argue that this could be due to the complexity of the task that the model cannot effectively learn. On the other hand, while the ``Edit'' task does not improve performance, it could help to make the model interpretable without using external tools.

Additionally, we study the case where we replace the GEC task with consecutive ``Edit'' and ``Apply'' tasks, 
% The motivation here was to 
decomposing GEC into a chain-of-thought process: first the model says what is wrong and what actions should be taken and then it applies these actions to the original sentence to produce the final output. Table~\ref{tab:task abl} shows that the model trained with only these two tasks severely underperforms, so ``classical'' GEC is still necessary.

\subsection{Auxiliary Tasks Analysis}

Next, we examine the model's performance on auxiliary tasks. For the analysis of ``Explain'' and ``Apply'' tasks, we use the model that had them both in pretraining; to analyze the ``Edit'' task, we use a model pretrained on all four tasks. We use the CoNLL-14 dataset for evaluation.

The ``Explain'' task urges the model to align to sentences and produce a correct sequence of edits to transform the source into the target. The exact match score of generated edits is 90\% compared to the gold standard edits, and the \fhalf score is $91.55$. Moreover, many ``errors'' here come from the ambiguity of how to perform replacements or from merging/splitting corrections compared to the gold standard. For example, for the original sentence ``{\it
People can also know much information about the celebrity in Twitter and Facebook such as Obama , Bill Gates and get the first-hand study materials on it .}'' an annotator suggested the following corrections: replace {\it know much} with {\it find out a great deal of}; delete {\it the}; replace {\it celebrity} with {\it celebrities}; replace {\it in} with {\it from}; replace {\it ,} with {\it and}; delete {\it the}; replace {\it it} with {\it these}. The model correctly predicts all of them except one. It splits the first correction into two: replace {\it know} with {\it find out} and replace {\it much} with {\it a great deal}.

The model deals quite well with the ``Apply'' task, getting 93.5\% accuracy. Most errors correspond to doing a different operation (e.g., inserting a new word instead of replacing one word with another), applying edits in the wrong place (e.g., placing new tokens after a wrong word), and simply ignoring some of the edits, especially if the list of edits is long. It appears that some errors arise from ambiguous descriptions of edits---in our approach we specify edits as, e.g., ``Insert the'', and the model has to decide where to insert it by itself---so performance could be improved further. For an example, consider the following sentence: {\it It is a concern that will be with us during our whole life , because we will never know when the ''potential bomb' ' will explode .} Here, the correction ``delete {\it will}'' targets the second {\it will}, but the model does not make changes to the original sentence, ignoring the correction. In this work, we restricted ourselves to simplistic prompts, and we leave an exploration of more detailed edit descriptions for further work.

The last auxiliary task, ``Edit'', is the prediction of a list of edits. This task is hard for the model: the exact match score is only 23.5\%, and the \fhalf score is 30.69. This low performance might be the reason why adding this task to the training pipeline decreases GEC performance. There are many cases where the model correctly applies an edit using a prediction prompt but omits it with the ``Edit'' prompt, which could indicate that the two tasks do not interact properly. It seems that the model struggles to predict a combination of connected or related corrections. For example, for the sentence ``{\it The people with albinism have sensitive skin and it needs regular treatment .}'', annotators gave the following edits: replace {\it The people} with {\it People}; replace {\it it} with {\it this}. But the model predicts the following: delete {\it The}; replace {\it it} with {\it they}. Thus, after correctly removing \emph{the} it does not capitalize {\it people}, and while replacing {\it it} with {\it they} it does not replace {\it needs} with {\it need}.

Since our model can be trained on both ``Edit'' and ``Apply'' tasks, it is tempting to apply a chain-of-thought procedure where we first predict the list of operations and then apply them to the original sentence. Applying such a procedure, we get the \fhalf score of $52.18$ on the CoNLL-14 dataset (precision $55.85$, recall $41.33$), i.e., the most problematic metric is precision (see also Table~\ref{tab:task abl}). The main problem of this task is the ambiguity of the target sentence and edits needed to obtain it: edits can interact with each other, which is hard to account for given only the source sentence. This chain-of-thought approach appears unnecessarily challenging, while adding the target sentence in the ``Explain'' task makes the problem much easier and more helpful for pretraining.

We have also studied how tasks interact with each other. In the first experiment, we correct a sentence with the model and then ask it to explain the corrections. We compare the explanations with edits produced by the ERRANT tool, obtaining an exact match score of 95.9\%, higher by 5.9\% than for human-corrected sentences. This indicates some interaction between tasks as the model's corrections are more explainable by itself. Second, we chain three tasks: first the model corrects a sentence, then it explains the corrections, and finally it applies edits to the original sentence. Interestingly, the exact match between the corrected sentence on the first step and the sentence obtained after chaining is 97.02\%, so the model is not always consistent across the tasks. Many errors come from the ambiguity of edits. It may be possible to use the discrepancy between corrected and generated sentences for filtering, either leaving the initial sentence intact or taking the intersection of the edits related to the two sentences. We find, however, that these approaches do not improve the final performance or even decrease it.

\subsection{Automatic Human Evaluation}
In this part, we compare the performance of our model with ChatGPT with chain-of-thoughts prompting, using the results by \citet{fang2023chatgpt} who showed that ChatGPT performs poorly on common benchmarks, far below state of the art fine-tuned models in the \fhalf metric, with high recall but very low precision. They show that the model often proposes corrections that are reasonable but far from minimal with respect to the number of required edits. To punish such over-corrections less, \citet{bryant-ng-2015-far} propose to use a test set where each erroneous sentence is annotated by several experts and averages the scores obtained in comparison to each annotator; this approach increases the scores for edits proposed by at least one annotator. Following \citet{fang2023chatgpt}, we use the evaluation set from this paper consisting of CoNLL-14 sentences with $10$ human annotations each. We also compare with a human-level baseline which is measured as the average score of each human annotator with respect to others. Table~\ref{tab: human eval} shows that even in this setup our model outperforms ChatGPT by a large margin; we compare with \citet{fang2023chatgpt} who use chain-of-thought prompting and report results in the zero-shot and few-shot ($1$, $3$, and $5$) settings.  Interestingly, ChatGPT performance is slightly below the human level, while our model performs significantly better. 
%Following this setup we calculate the
% score of each annotator by comparing its
% corrections to the other 9 annotators. The average
% of the scores is considered as the final score
% for the human-level performance. Our result is 81.358 if we do just average of 10.

\begin{table}[!t]
\centering\setlength{\tabcolsep}{10pt}\small
% \begin{threeparttable}
\begin{tabular}{cc}
\toprule
\textbf{System} & \fhalfb \\
\midrule
Human level    & 72.58   \\
\midrule
Transformer & 66.97 \\
GECToR & 80.49 \\
T5-large & 81.19 \\
\midrule
ChatGPT (0-shot)   & 69.74 \\
ChatGPT (1-shot)   & 71.55 \\
ChatGPT (3-shot)   & 71.73 \\
ChatGPT (5-shot)   & 70.66 \\
\midrule
Our & \textbf{81.36} \\
\bottomrule
\end{tabular}
% \end{threeparttable}
\caption{Comparison to automatic human evaluation performance on 10 references for CoNLL-14; ChatGPT results are taken from \citet{fang2023chatgpt}.}
\label{tab: human eval}
\end{table}

\section{Conclusion}\label{sec:concl}

In this work, we have introduced a new approach to training and fine-tuning sequence-to-sequence models for the grammatical error correction task based on multi-task pretraining and optimized training schedules. The proposed approach fares better than previous state of the art models with a much smaller model size and outperforms models of comparable size by a wide margin on both CoNLL-14 and BEA-test datasets; in particular, we have achieved results exceeding state of the art models based on T5-XXL (11B parameters) with a BART-based model (400M parameters). Our multi-task approach encodes auxiliary tasks as text rewriting problems and does not require any changes in the model architecture. We believe that the proposed approach is of significant value for practical GEC pipelines and see several directions for possible improvements in further work, including modifications of text encodings for auxiliary tasks and adding new auxiliary tasks to the training process.

\section{Limitations}\label{sec:limits}

Limitations of our study can serve as motivation for further work. First, we do not test our approach in the increasingly important multilingual setup; we believe that our ideas may contribute even more to multilingual models that need greater robustness. 
Second, the highest possible results in GEC are currently obtained by ensembles rather than individual models, including the recent state of the art approach by~\citet{qorib-etal-2022-frustratingly}; while ensembling would make our reduced model size less important, it would be interesting to study how well our model can perform in such ensembles.
Third, we consider only simple prompts and small models, and our results could be extended further.

\section*{Acknowledgements}

The work of Sergey Nikolenko was done under support of the grant no.~075-15-2022-289 for the creation and development of Euler International Mathematical Institute.

% \section*{Societal Impact}

\bibliography{anthology,custom}
\bibliographystyle{acl_natbib}

% \documentclass[11pt]{article}

% \usepackage[final]{EMNLP2023}

% % Standard package includes
% \usepackage{times}
% \usepackage{latexsym}
% \usepackage{paralist}
% \usepackage{enumerate}
% \usepackage{numprint}
% \usepackage{xspace}

% % For proper rendering and hyphenation of words containing Latin characters (including in bib files)
% \usepackage[T1]{fontenc}
% % For Vietnamese characters
% % \usepackage[T5]{fontenc}
% % See https://www.latex-project.org/help/documentation/encguide.pdf for other character sets

% % This assumes your files are encoded as UTF8
% \usepackage[utf8]{inputenc}
% \usepackage{hyperref}

% % This is not strictly necessary, and may be commented out,
% % but it will improve the layout of the manuscript,
% % and will typically save some space.
% \usepackage{microtype}
% \usepackage{booktabs}
% \usepackage{threeparttable}
% \usepackage{multirow}
% \usepackage{graphicx}
% \usepackage{xspace}

% If the title and author information does not fit in the area allocated, uncomment the following
%
%\setlength\titlebox{<dim>}
%
% and set <dim> to something 5cm or larger.

\newpage

\def\src{\mathrm{src}}
\def\tgt{\mathrm{tgt}}
\def\hyp{\mathrm{hyp}}
\def\wi{{W\&I+L}\xspace}

\def\fhalf{\ensuremath{\mathrm{F}_{0.5}}\xspace}
\def\fhalfb{\ensuremath{\mathbf{F}_{0.5}}\xspace}

% \begin{document}
\appendix
\setcounter{secnumdepth}{1}

\clearpage

\section{Error Type Distribution in Datasets}
In order to analyze the difference between the GEC datasets, we compute the error type statistics using the ERRANT tool. Next, we compare each dataset with W\&I+L-dev, see Figures~\ref{fig:wi_dev_wi_train_comp}-\ref{fig:wi_dev_bea_train_comp}. We see that \wi-train and \wi-dev have only a slight discrepancy in the proportions of each error type, with major differences coming from punctuation and spelling errors. Hence, we should expect that the distribution on \wi-test is close but not exactly the same as the articles that comprise the datasets were written by different people. Thus, extensive hyperparameter search on \wi-dev may lead to lower performance. The FCE dataset is also close to these datasets except for SPELL, NOUN, and VERB errors. These differences can be explained by the fact that essays were written by authors who had only begun learning English, and these error types are more common for them.

Comparing \wi-dev with NUCLE and Lang-8 datasets, we note that for some error types the difference is striking. We highlight that the proportion of OTHER errors is high for both datasets. In the ERRANT toolkit, OTHER corresponds to errors that cannot be labeled by a predefined set of error types. After examining some instances with that error type, we have noticed that a large part of them is related to sentence rephrasing or rewriting, perhaps regrouping some parts of the sentence. Training on such examples might hurt the model's performance because it would becomes more aggressive in performing corrections, leading to much lower precision. Therefore, we mark those datasets as ``noisy'' and others as ``clean''.

The combination of all datasets---FCE, \wi-train, Lang-8, and NUCLE---does not improve the situation, as shown in Figure~\ref{fig:wi_dev_bea_train_comp}. Nevertheless, if we train only on FCE and \wi-train, we obtain poor performance. This indicates that not only the distribution of errors should be close but also the errors should be diverse in order to generalize well. These two factors reveal why we needed multi-stage fine-tuning.

Our approach can be viewed as curriculum learning where we gradually train a model on ``less noisy'' data.

\def\mywid{.85\linewidth}
\def\myvspace{\vspace{-.5cm}}

\begin{figure*}[!t]
\centering
\includegraphics[width=\mywid]{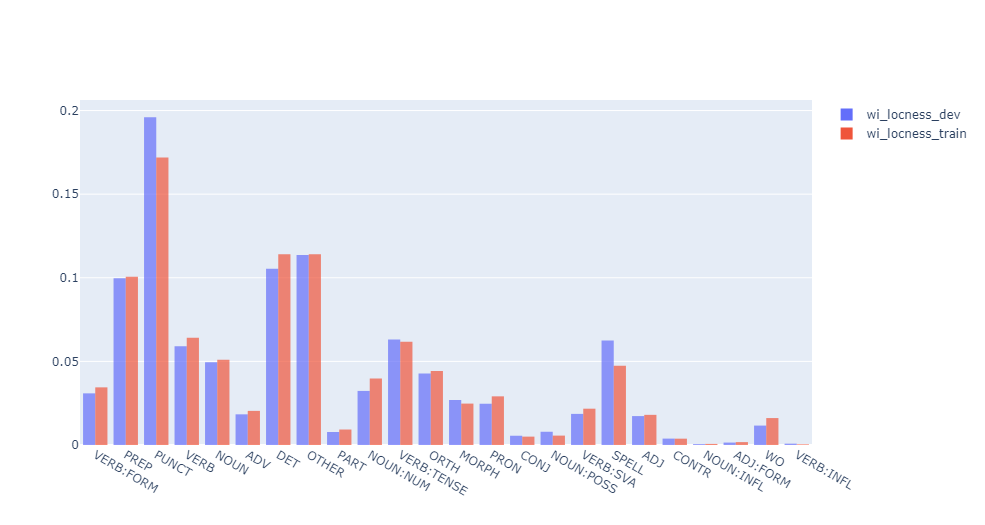}\myvspace
\caption{\wi-dev and \wi-train error type comparison.}
\label{fig:wi_dev_wi_train_comp}
% \end{figure*}

% \begin{figure*}[!t]
\centering
\includegraphics[width=\mywid]{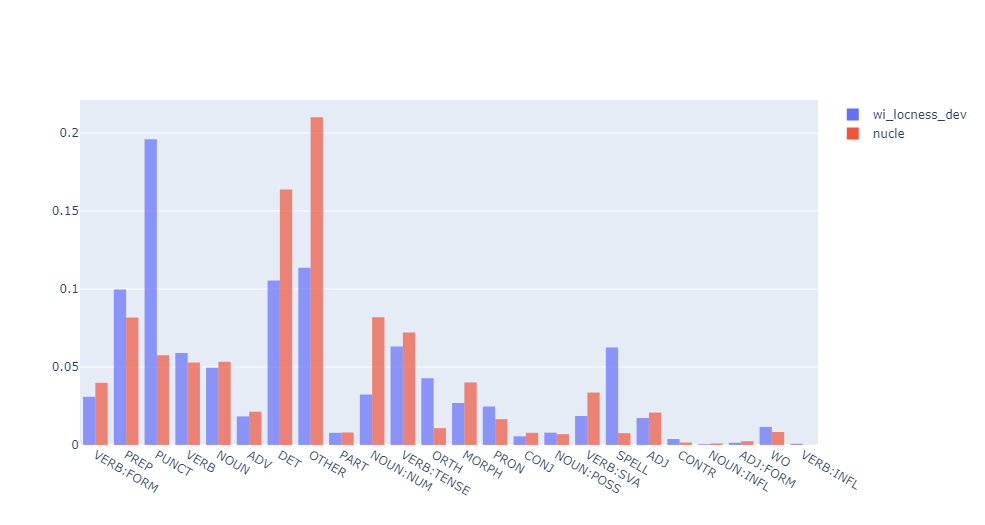}\myvspace
\caption{\wi-dev and NUCLE error type comparison.}
\label{fig:wi_dev_wi_train_comp}
% \end{figure*}

% \begin{figure*}[!t]
\centering
\includegraphics[width=\mywid]{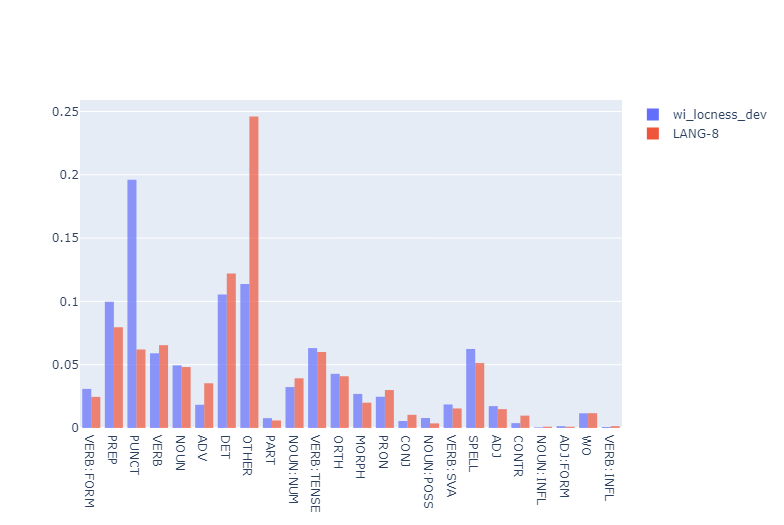}\myvspace
\caption{\wi-dev and LANG-8 error type comparison.}
\label{fig:wi_dev_lang8_comp}
\end{figure*}

\begin{figure*}[!t]
\centering
\includegraphics[width=\mywid]{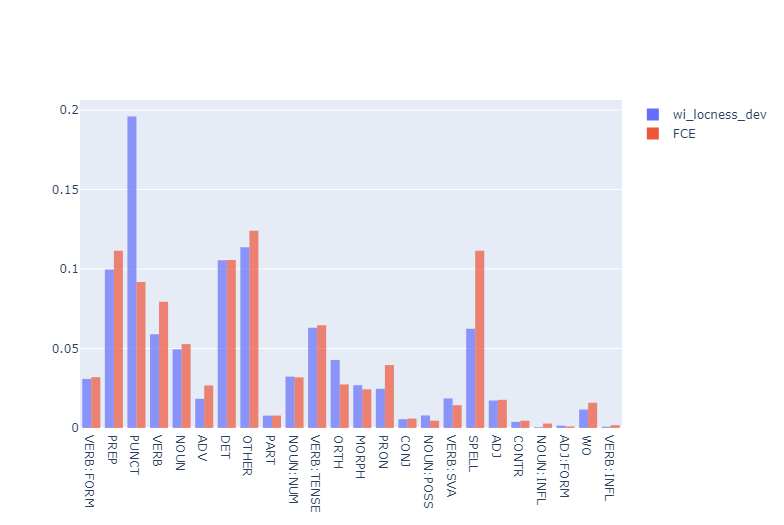}\myvspace
\caption{\wi-dev and FCE error type comparison.}
\label{fig:wi_dev_fce_comp}
% \end{figure*}

% \begin{figure*}[!t]
\centering
\includegraphics[width=\mywid]{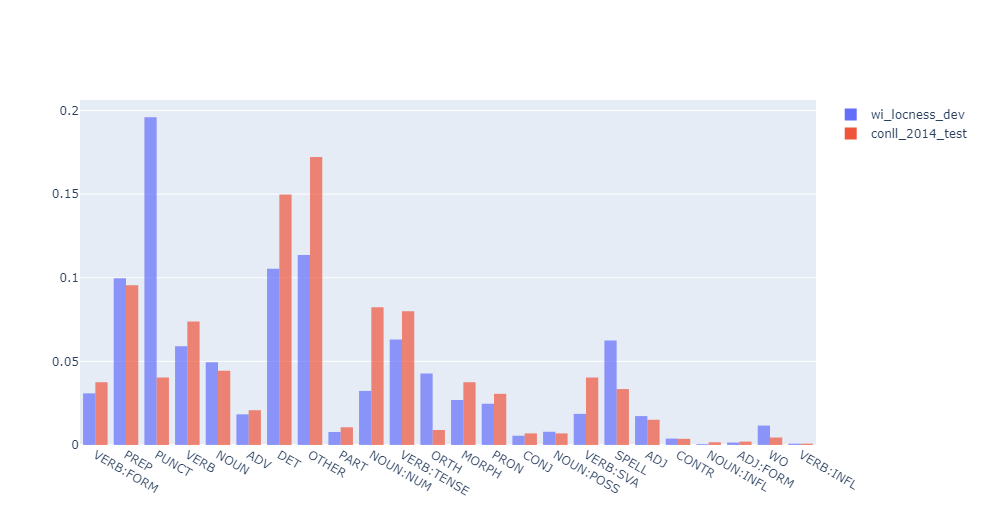}\myvspace
\caption{\wi-dev and CoNLL-14 error type comparison.}
\label{fig:wi_dev_conll14_comp}
% \end{figure*}

% \begin{figure*}[!t]
\centering
\includegraphics[width=\mywid]{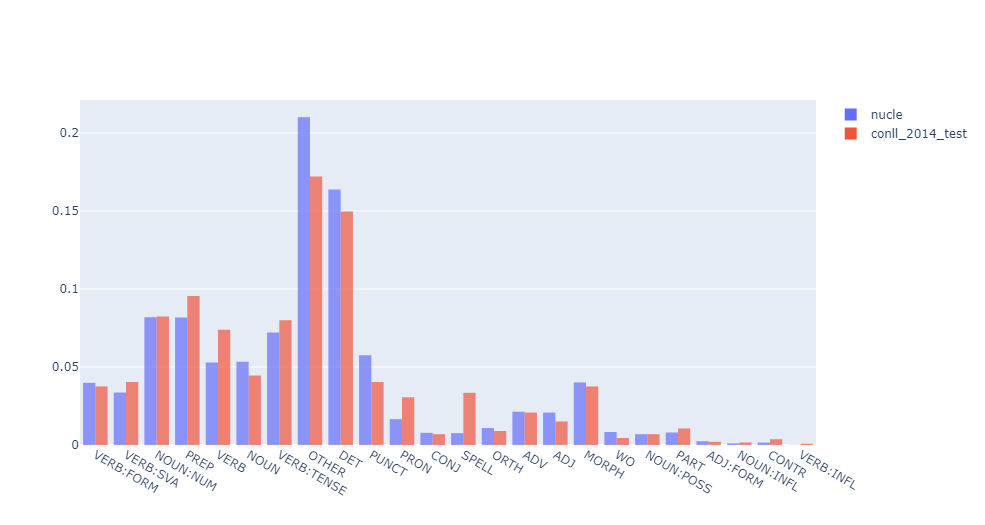}\myvspace
\caption{NUCLE and CoNLL-14 error type comparison.}
\label{fig:nucle_conll14_comp}
\end{figure*}

\begin{figure*}[!t]
\centering
\includegraphics[width=\mywid]{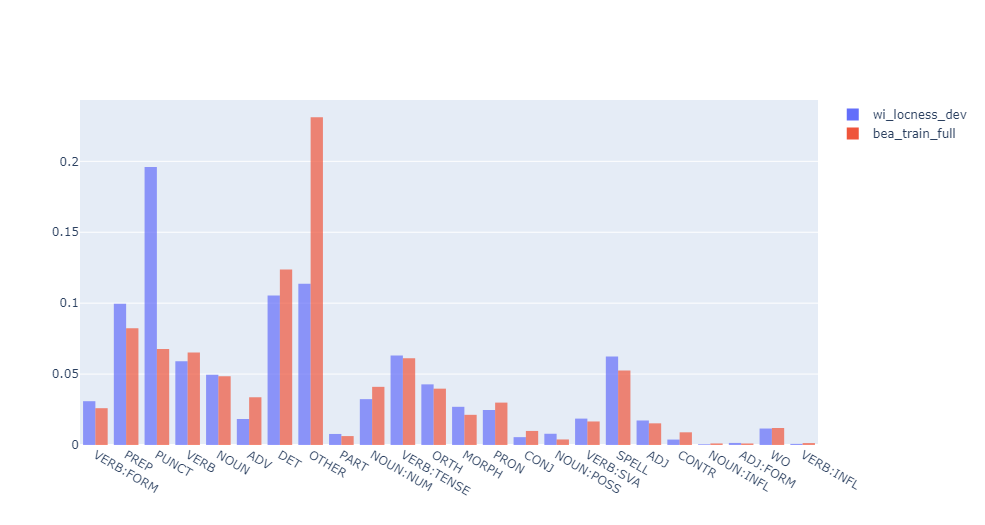}\myvspace
\caption{\wi-dev and BEA-train error type comparison.}
\label{fig:wi_dev_bea_train_comp}
\end{figure*}

\section{Model comparison}
Performance on the \wi-dev dataset of the 3-stage model and of the multi-task model with the improved training schedule seems to be marginal: 62.62 versus 62.67. However, the models' behavior on \wi-dev differs. The 3-stage model has higher recall and the other has higher precision. As we have noted in the previous section, error distributions on parts of \wi datasets differ from each other since different people are prone to making different types of errors. Therefore, we expect that a model with higher precision that makes more accurate corrections would generalize better. Looking at the models' performance on \wi-test and CoNLL-14, we see that the gap between the models is indeed large. 

To draw the distinction further, we compute the statistics of error types corrected and introduced by them with the ERRANT toolkit. Figure~\ref{fig:model_comp_absolute} presents the absolute number of corrected, generated, and non-corrected errors for every type. Again, we see that the 3-stage model is more aggressive. It corrects more errors of each type but also introduces more new errors. In Figures~\ref{fig:model_comp_corr_prop}-\ref{fig:model_comp_left_prop}, we show the distribution of corrected, induced, and non-corrected errors by type. Again, we see that the models' behavior differs.

\renewcommand{\mywid}{.8\linewidth}
\renewcommand{\myvspace}{\vspace{-.2cm}}

\begin{figure*}[!t]
\centering
\includegraphics[width=\mywid]{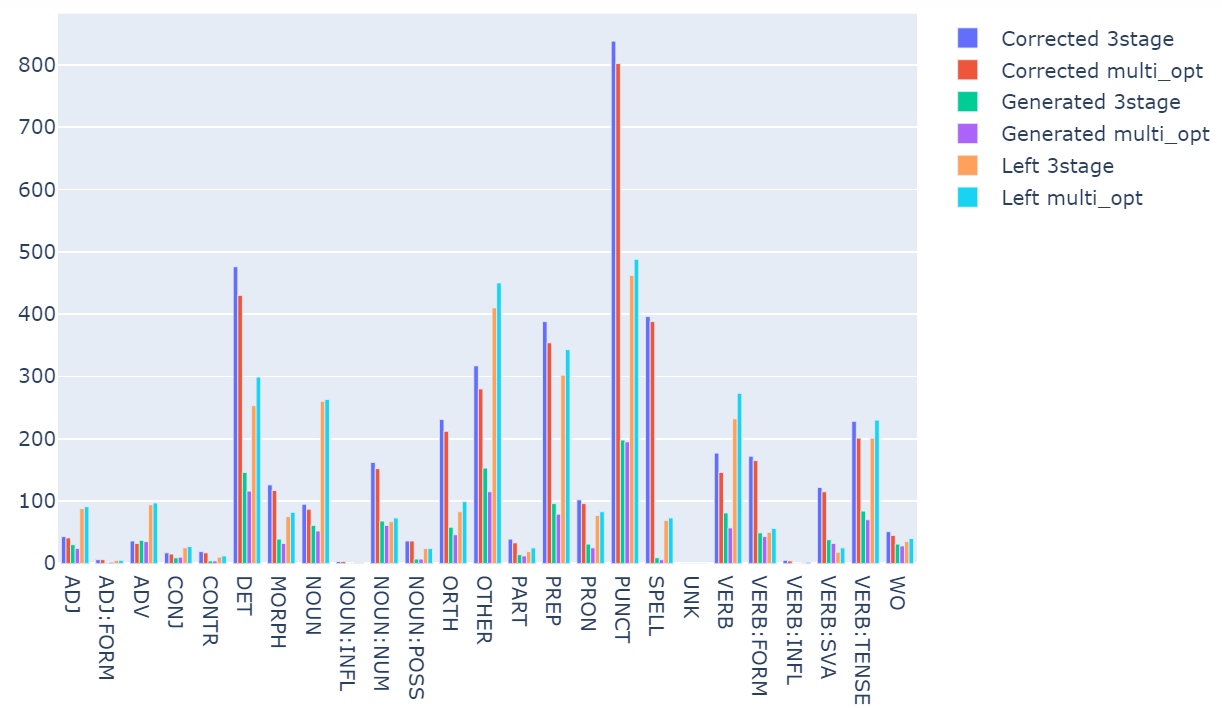}\myvspace
\caption{3-stage and multi-task improved schedule models comparison. Corrected --- errors corrected by the model. Generated --- errors introduced by the model. Left --- errors that were not corrected.}
\label{fig:model_comp_absolute}
\end{figure*}

\begin{figure*}[!t]
\centering
\includegraphics[width=\mywid]{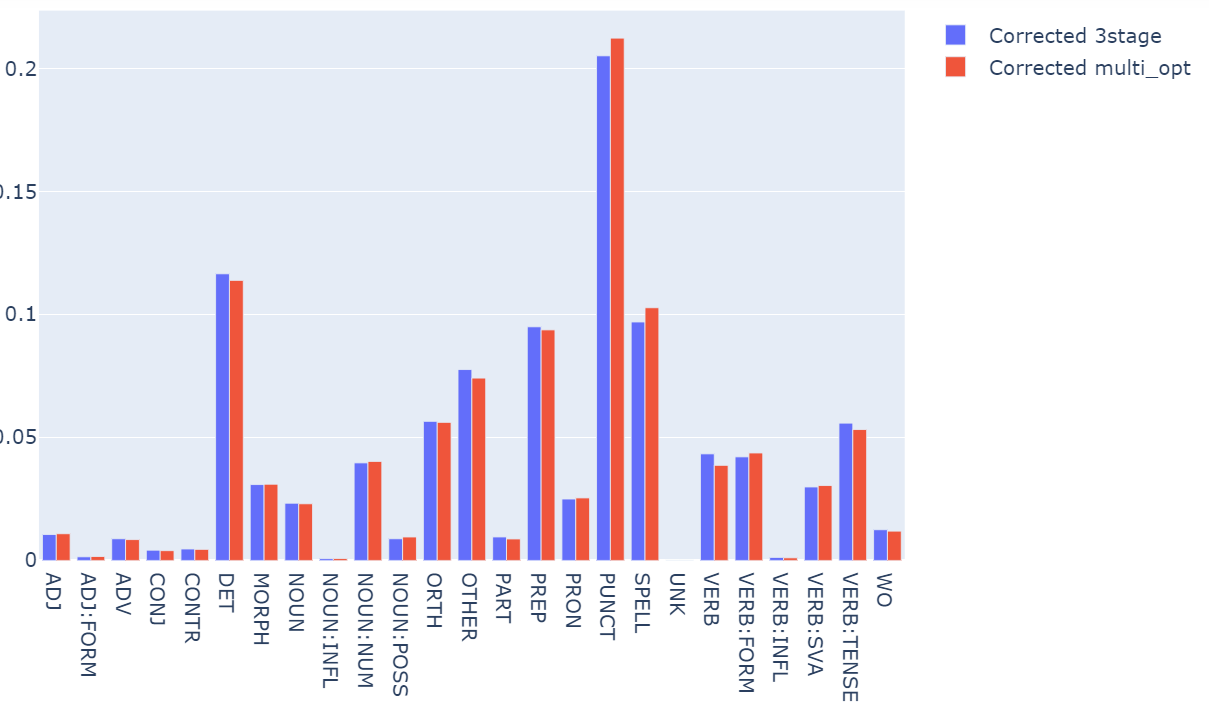}\myvspace
\caption{3-stage and multi-task improved schedule models corrected errors distribution by type.}
\label{fig:model_comp_corr_prop}
% \end{figure*}

% \begin{figure*}[!t]
\centering
\includegraphics[width=\mywid]{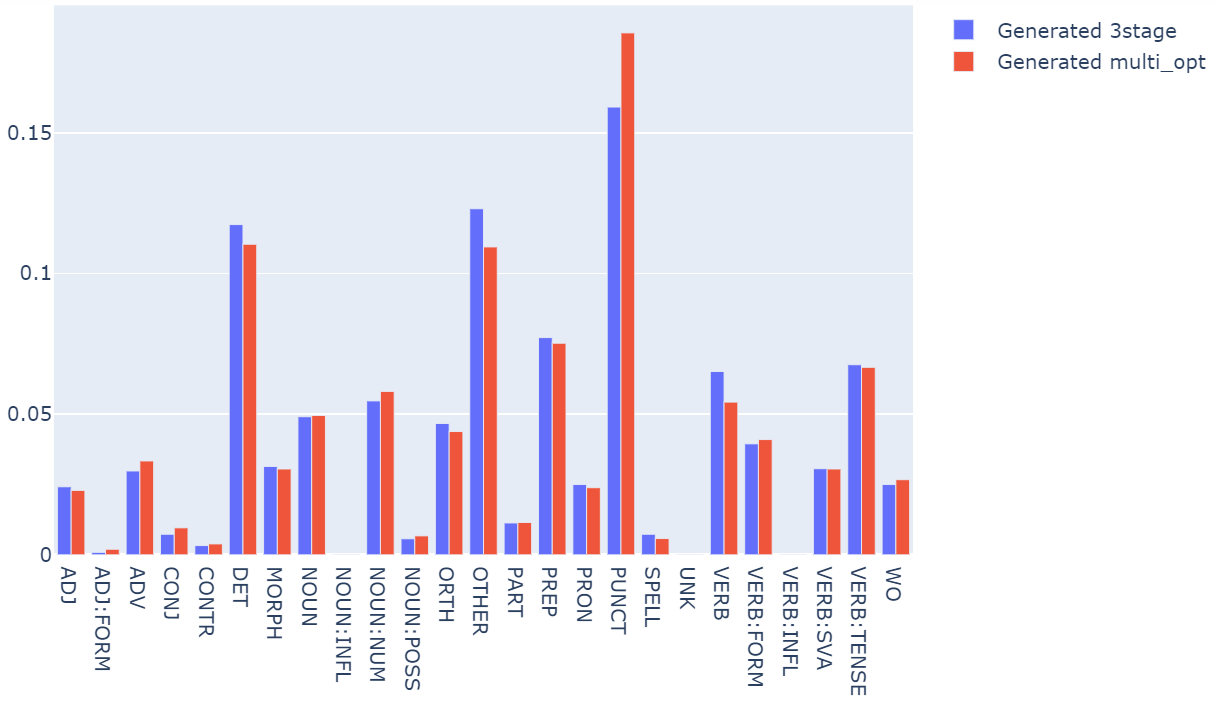}\myvspace
\caption{3-stage and multi-task improved schedule models introduced errors distribution by type.}
\label{fig:model_comp_gen_prop}
% \end{figure*}

% \begin{figure*}[!t]
\centering
\includegraphics[width=\mywid]{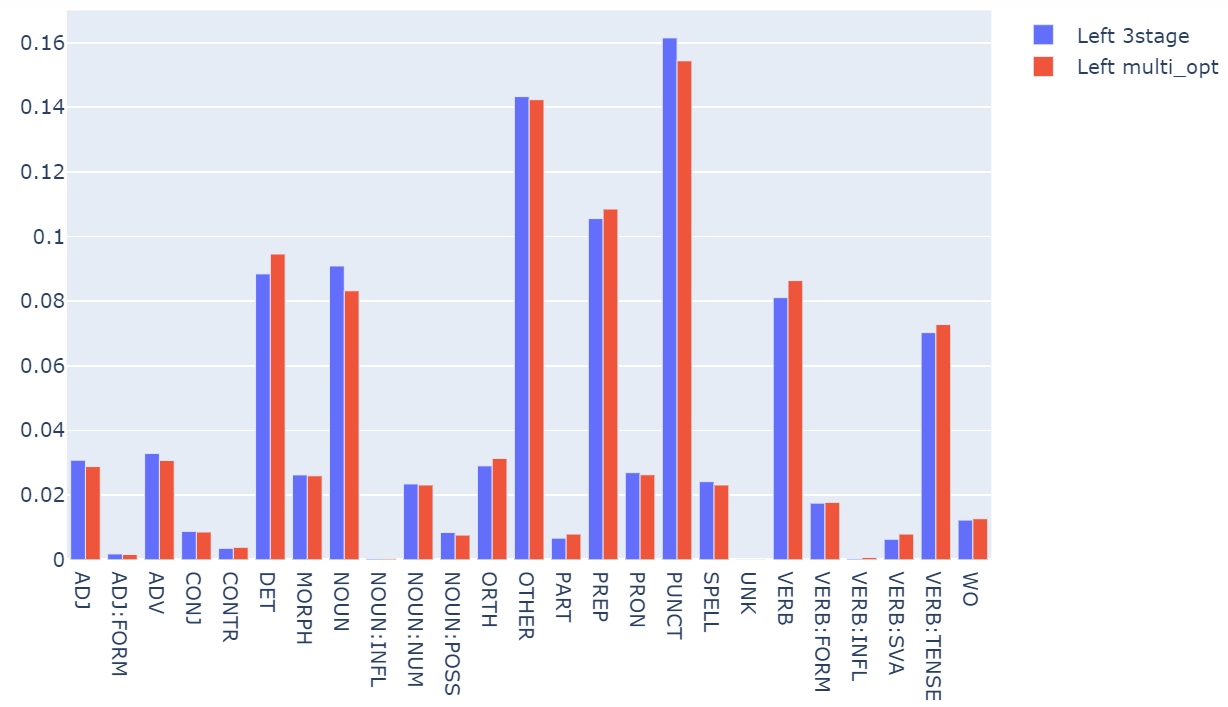}\myvspace
\caption{3-stage and multi-task improved schedule models non-corrected errors distribution by type.}
\label{fig:model_comp_left_prop}
\end{figure*}

%\bibliography{references.bib}
% \end{document}

\end{document}